\documentclass[runningheads]{llncs}
\usepackage{amssymb}
\usepackage{graphicx}
\usepackage{amsmath}
\usepackage{bm}
\usepackage[colorlinks,
            linkcolor=red,       
            anchorcolor=red,  
            citecolor=green,        
            ]{hyperref}

\begin{document}
\title{AG-NeRF: Attention-guided Neural Radiance Fields for Multi-height Large-scale Outdoor Scene Rendering}

\author{Jingfeng Guo, Xiaohan Zhang, Baozhu Zhao, Qi Liu}
\authorrunning{Guo, Jingfeng, et al.}
\institute{South China University of Technology}

\maketitle              

\begin{abstract}
Existing neural radiance fields (NeRF)-based novel view synthesis methods for large-scale outdoor scenes are mainly built on a single altitude. Moreover, they often require \textit{a priori} camera shooting height and scene scope, leading to inefficient and impractical applications when camera altitude changes. In this work, we propose an end-to-end framework, termed AG-NeRF, and seek to reduce the training cost of building good reconstructions by synthesizing free-viewpoint images based on varying altitudes of scenes. Specifically, to tackle the detail variation problem from low altitude (drone-level) to high altitude (satellite-level), a source image selection method and an attention-based feature fusion approach are developed to extract and fuse the most relevant features of target view from multi-height images for high-fidelity rendering. Extensive experiments demonstrate that AG-NeRF achieves SOTA performance on $56$ Leonard and Transamerica benchmarks and only requires a half hour of training time to reach the competitive PSNR as compared to the latest BungeeNeRF.
\end{abstract}
\begin{keywords}
Novel View Synthesis, NeRF, Large-scale Outdoor Scene Rendering
\end{keywords}
\section{Introduction}
\begin{figure}[t]
    \centering
    \includegraphics[width=\textwidth]{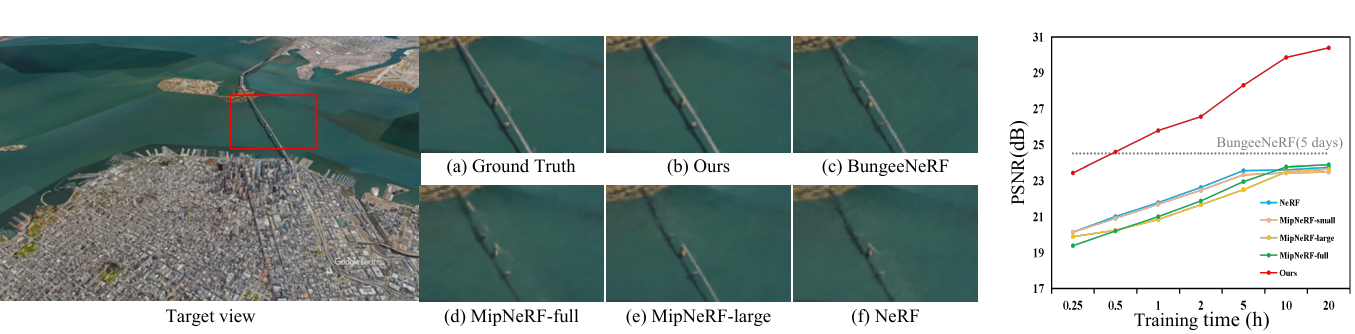}
           \caption{Performance comparisons on two benchmark datasets. \textbf{Left}: visualization on Transamerica dataset. The visual results show that the proposal outperforms other competitors and can reconstruct the bridge completely. \textbf{Right}: PSNR versus training time on $56$ Leonard dataset. Compared with others, we observe that ours gets $ 6\sim7 $ dB improvement at PSNR. Moreover, it is worth noting that the proposed method only requires a half hour of training on a single RTX $4090$ GPU to achieve competitive performance as the latest BungeeNeRF \cite{xiangli2022bungeenerf} (training over five days).}
           \label{fig: Qualitative comparisons on Transamerica dataset and traintime}
\end{figure}

Large-scale outdoor scene reconstruction has an important application prospect to digitize a smart city in virtual reality and augmented reality. With the advance of neural radiance fields (NeRF) \cite{nerf}, the success spurs numerous researchers to study NeRF with high-frequency positional encoding for single object scene reconstruction and novel view synthesis, and achieves impressive results. Nevertheless, due to the limited model capacity, NeRF-based variants can only represent scenes reasonably at a macro scale yet exhibit excessively blurry artifacts and incomplete reconstruction when navigating closer to inspect micro details for large-scale outdoor scenes, as shown on the left of Fig. \ref{fig: Qualitative comparisons on Transamerica dataset and traintime}. To address that, several approaches \cite{tancik2022block,turki2022mega,zhenxing2022switch} geographically decomposed the scene into several cells and trained a sub-NeRF for each cell before merging them, while others applied plane and grid features in parallel with positional encoding to achieve efficient modeling \cite{xu2023grid,zhang2023efficient}. \textit{However, all of them reconstruct large-scale scenes at the basis of identical low-altitude drone photos, rendering them for dealing with images of the same scene captured at varying heights in vain. This limitation arises from the significant detail variation during camera altitude changes. For the same scene, images captured at high-altitude primarily consist of low-frequency details, whereas images taken at low-altitude include more high-frequency details.} The pioneer to address large-scale outdoor scene reconstruction at varying heights was BungeeNeRF \cite{xiangli2022bungeenerf}, by applying a progressive growth network with residual block and a multi-stage training paradigm to learn a hierarchy of scene representations. Nonetheless, to activate high-frequency channels in NeRF’s positional encoding, BungeeNeRF \cite{xiangli2022bungeenerf} has to take use of the camera height to accurately partition the training dataset in a hand-crafted way. Furthermore, as the height increases, it inevitably grapples with an extensive model capacity, which takes up expensive training time non-amicable to limited GPU computing resources.

We seek to reduce the training cost of building good reconstructions by synthesizing novel views of large-scale scenes captured at different levels. To that end, we design the source
image selection and attention-based feature fusion approaches to extract potential features from images at different heights as scene priors for NeRF processing. The proposed AG-NeRF significantly outperforms the baseline BungeeNeRF \cite{xiangli2022bungeenerf} and, as corroborated by indicative empirical results in Fig.~\ref{fig: Qualitative comparisons on Transamerica dataset and traintime}, competitive to the state-of-the-art in terms of accuracy and speed. Our work makes notable contributions summarized as follows:

\begin{itemize}
    \item We propose an end-to-end novel view synthesis framework, called AG-NeRF, for large-scale outdoor scene reconstruction. Different from the existing novel view synthesis approaches using drone photos of the same height, the proposed AG-NeRF is not affected by this limitation and is applicable to images captured at different levels. Moreover, the camera height is not prior to know.
    \item The proposed framework is helpful in providing the most relevant features for synthesizing the target view, and enabling high-quality image rendering across all heights. This has been demonstrated by the comparison with the SOTA BungeeNeRF \cite{xiangli2022bungeenerf}, where our approach achieves almost $6$ dB increase in PSNR.
    \item Compared to the BungeeNeRF \cite{xiangli2022bungeenerf}, ours takes up only a half hour of training on a single RTX $4090$ GPU to achieve remarkable performance, while the competitor requires $5$ days for training.
\end{itemize}

\section{Related work}

\subsection{NeRF and Its extension}
NeRF \cite{nerf} has been extensively used in $3$D reconstruction and novel view synthesis owing to its detailed scene geometry representation with complex occlusions, where views are synthesized by querying $3$D point coordinates along camera rays and volume rendering is utilized to project the output colors and densities into an image. Recent works have been proposed to extend NeRF to unbounded scenes \cite{zhang2020nerf++,martin2021nerf,barron2022mip}, dynamic scenes \cite{pumarola2021d,lin2022efficient,Li_2023_CVPR,jiang2023alignerf}, few-shot setting \cite{yu2021pixelnerf,yang2023freenerf,roessle2022dense,deng2022depth,yuan2022neural}, and large-scale outdoor scenes \cite{tancik2022block,turki2022mega,zhenxing2022switch,xu2023grid,zhang2023efficient}. Another line of works uses display representations \cite{muller2022instant,chen2022tensorf,sun2022direct,liu2020neural,fridovich2022plenoxels,kerbl3Dgaussians} to accelerate NeRF convergence and inference processes, achieving a huge speed increase.

\subsection{Large-scale Outdoor Scene Reconstruction and Rendering}
Much effort has been devoted to extending NeRF to address large-scale outdoor scene reconstruction and rendering. Approaches like Block-NeRF \cite{tancik2022block} partition streets into discrete blocks and train a sub-NeRF for each block. Similarly, Mega-NeRF \cite{turki2022mega} divides drone-captured scenes into separate cells and trains a sub-NeRF for each cell. Switch-NeRF \cite{zhenxing2022switch} points out that manually crafted scene decomposition relies on prior knowledge of the target scene, which limits the generalized use of these models, so a learnable gated network is designed to dynamically allocate 3D points to different sub-NeRF networks. More recently, Grid-NeRF \cite{xu2023grid} combines NeRF-based methods and grid-based approaches, jointly optimizing two branches for scene reconstruction. GP-NeRF \cite{zhang2023efficient} integrates orthogonal 2D high-resolution tri-plane feature and 3D hash-grid feature to achieve efficient scene representation. These methods achieve a fantastic novel view synthesis performance at the same height as the drone view. However, real-world scenes often involve images captured at varying heights even with substantial height disparities, which makes the rendering quality poor. BungeeNeRF \cite{xiangli2022bungeenerf} was proposed to employ a progressive training paradigm with residual block, dynamically expanding the neural network and synchronously adjusting the training images as the camera height increases. However, the BungeeNeRF requires an explicit split of scales among the input images, which has to proceed with manual adjustment. Moreover, the network becomes wider or deeper along with the growth of parameters as the camera height increases, leading to multi-stage training for several days. In contrast, the proposal is fully differentiable and can therefore be trained end-to-end. Different with BungeeNeRF, we directly extract more favorable scene priors from images captured at different heights. This enables us to utilize a tiny MLP for color production. Consequently, no manual intervention operation is required and only a half hour of training time is taken up to achieve competitive performance as BungeeNeRF.

\begin{figure}[t]
	\centering
	\includegraphics[width=\textwidth]{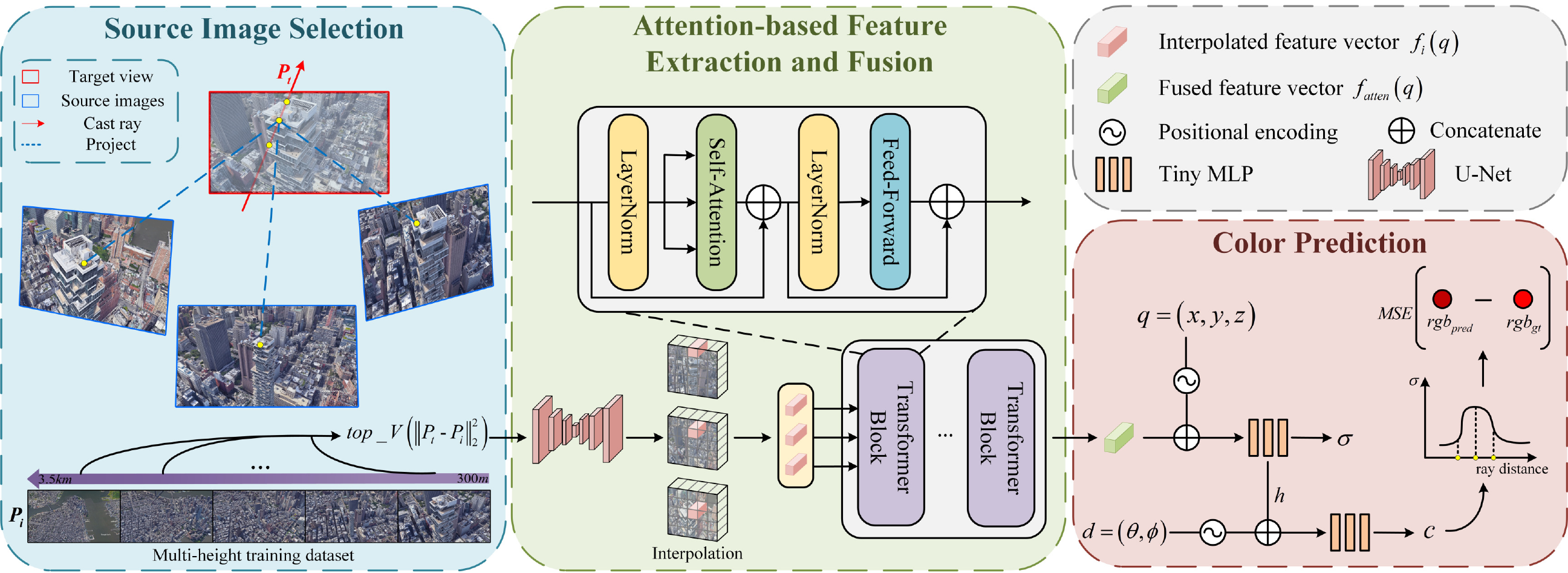}
	\caption{\textbf{Our pipline.} First, according to the camera's external matrix, we select source images that are most similar to the target view from different heights. Next, a trainable U-Net-like network extracts feature maps from these source images. The $3$D sample points along the rays are then projected back onto the image planes and interpolated for the corresponding feature vectors. Subsequently, these feature vectors interact with each other through an attention-based feature fusion approach and are fed into MLPs along with positional encoding. Finally, pixel color is calculated by volume rendering.}
	\label{fig:pipline}
\end{figure}
 
\section{Method}
\textbf{Overview.} Given a set of multi-height training data, our goal is to synthesize images from any view at any altitude. We obtain camera intrinsic and extrinsic matrices through COLMAP \cite{schonberger2016structure} or Google Earth Studio. As shown in Fig. \ref{fig:pipline}, for each target view, we select $V$ source images via our source image selection method (Sec. \ref{Source images selection}). For each sampled point along casted rays, we project them back to the source images and obtain fused features through our attention-based feature extraction and fusion approach (Sec. \ref{Attention-based feature extraction and fusion}). Finally, colors and densities along each ray through volume rendering are composited to produce a synthesized image (Sec. \ref{Rendering and training}).

\subsection{Source Image Selection} \label{Source images selection}
Current NeRF-based large-scale outdoor reconstructors encode the entire scene into MLP layers to enhance the scene representations, which comes at the cost of high computational complexity. Different with them, we select few images from different heights by scoring to obtain scene priors. This is because the regions captured by close cameras constitute a subspace of those captured by remote cameras. Furthermore, these priors act as a guide to activate the high-frequency channels of positional encoding in close views, or to activate low-frequency channels in distant views. This adaptive mechanism effectively addresses the challenge posed by detail variation caused by altitude changes.

Given a target view with camera extrinsic matrix ${P_t} \in {{\mathbb R}^{3 \times 4}}$, we leverage the extrinsic matrix distance between two views as a metric to measure the similarity of two views, and select $V$ images that are the closest to the target view at different heights as the source images $\left\{ {{S_i}\left| {i = 1, \ldots ,V} \right.} \right\}$. By doing so, the selected source images overlap highly with the target view. This process can be written as:
\begin{equation}
\label{con:source image selection}
source\_imgs = top\_V\left( {\left\| {{P_t} - {P_i}} \right\|_2^2\left| {i = 1, \ldots ,M} \right.} \right)
\end{equation}
where $P_i$ represents the camera's external matrix for the $i$-th image in the training dataset, $M$ is the number of images in the training dataset,  $top\_V$ means to find the first $V$ minimum values, and ${\left\| {} \right\|_2}$ denotes the ${l_2}$ norm.

\subsection{Attention-based Feature Extraction and Fusion} \label{Attention-based feature extraction and fusion}
\textbf{Feature Extraction.} We apply a trainable U-Net-like network to extract feature maps $\left\{ {{F_i}\left| {i = 1, \ldots ,V} \right.} \right\}$ from the source images. Subsequently, for a sampled point $q \in {{\mathbb R}^3}$ along a ray, we project $q$ onto the image planes using \textit{a priori} camera intrinsic matrices, then bilinear interpolation is applied between pixel-wise features to extract feature vectors. We also sample RGB colors on the images and concatenate them to the extracted feature vectors to form the final feature vectors $\left\{ {{f_i}\left( q \right)\left| {i = 1, \ldots ,V} \right.} \right\}$. The feature extraction step can be mathematically expressed as:
\begin{equation}
{f_i}\left( q \right) = Inter({S_i},{\pi _i}\left( q \right)) \oplus Inter({F_i},{\pi _i}\left( q \right))
\end{equation}
where $ \oplus $ denotes concatenation in channel dimension, ${\pi _i}\left( q \right)$ represents the coordinates on the image plane ${{S_i}}$ obtained by projecting 3D point $q$.

\textbf{Feature Fusion.} While these source images with high overlap or panoramic do provide priors for the target view, they also introduce irrelevant information that is absent in the target view. Hence, an attention-based feature fusion approach is developed, which combines all the feature vectors, to maximize the relevance between the fused feature and the target pixel. It is formalized as:
\begin{equation}
{f_{atten}}\left( q \right) = Transformer\left( {{f_1}\left( q \right),...,{f_V}\left( q \right)} \right)
\end{equation}
It is worth noting that the used transformer costs less computation as we only pay attention to the feature vectors extracted by interpolation.

\subsection{Rendering and Training} \label{Rendering and training}
\textbf{Point Sampling.} Similar with the NeRF \cite{nerf}, a hierarchical sampling approach is applied. A coarse and a fine networks are simultaneously optimized. We first uniformly sample ${N_c}$ points to obtain the output of the coarse network, and then produce a more informed samples along each ray where samples are biased towards the surface of the object. Similarly, a set of ${N_f}$ points is sampled and all ${N_c} + {N_f}$ points are applied to render fine results.

\textbf{Training Objective.} Positional encoding is employed to map 3D point coordinates ${q_k} = \left( {x,y,z} \right)$ and directions ${d_k} = \left( {\theta ,\phi } \right)$ into high-dimensional space, making our MLP easier to approximate higher-frequency functions. We concatenate the positional encoding $\left( {\gamma \left( q_k \right),\gamma \left( d_k \right)} \right)$ and the fused features ${{f_{atten}}\left( q_k \right)}$ as input to the MLPs to obtain the color ${{c_k}}$ and density ${{\sigma _k}}$ of the $k$-th sample on the ray. That is:
\begin{equation}
\begin{aligned}
{\sigma _k},{h_k} &= ML{P_1}\left( {\gamma \left( {{q_k}} \right),{f_{atten}}\left( {{q_k}} \right)} \right) \\
{c_k} &= ML{P_2}\left( {\gamma \left( {{d_k}} \right),{h_k}} \right)
\end{aligned}
\end{equation}

The volume rendering is then used to get the predicted color $\widehat C\left( r \right)$ of each pixel:
\begin{equation}
\widehat C\left( r \right) = \sum\limits_{k = 1}^N {{T_k}\left( {1 - \exp \left( { - {\sigma _k}{\delta _k}} \right)} \right){c_k}} 
\end{equation}
where ${{\delta _k}}$ is the distance between adjacent sample points, ${T_k} = \exp \left( { - \sum\limits_{j = 1}^{k - 1} {{\sigma _k}{\delta _k}} } \right)$ is the accumulated transmittance, and $N$ is the number of sample points along a cast ray $r$.

We render the color of each ray using both the coarse and fine set of samples, and minimize the total squared error between the rendered colors and true pixel colors for training:
\begin{equation}
{\cal L} = \sum\limits_{r \in R} {\left[ {\left\| {C\left( r \right) - {{\widehat C}_c}\left( r \right)} \right\|_2^2 + \left\| {C\left( r \right) - {{\widehat C}_f}\left( r \right)} \right\|_2^2} \right]} 
\end{equation}
where $R$ is a set of rays in each batch, ${C\left( r \right)}$, ${{{\widehat C}_c}\left( r \right)}$ and ${{{\widehat C}_f}\left( r \right)}$ are the ground truth, coarse predicted, and fine predicted colors for ray $r$, respectively. 

\section{Experiments}
\subsection{Implementation Details}
\textbf{Datasets.}
Following BungeeNeRF \cite{xiangli2022bungeenerf}, we evaluate our AG-NeRF on two large-scale outdoor scene datasets: $56$ Leonard and Transamerica. These datasets comprehensively depict the real-world landscapes of New York and San Francisco, spanning a diverse range of altitudes from drone-level (about $300m$) to satellite-level (about $3.5km$). The images and camera parameter matrices for both datasets are obtained from Google Earth Studio by BungeeNeRF \cite{xiangli2022bungeenerf}. 

\textbf{Evaluation Metrics.}
Similar with the NeRF-based approaches, we test our method against competitors in quantitative metrics as PSNR, SSIM \cite{wang2004image}, and LPIPS \cite{zhang2018unreasonable} implemented with VGG.

\begin{figure}[!h]
    \centering
    \includegraphics[width=\textwidth]{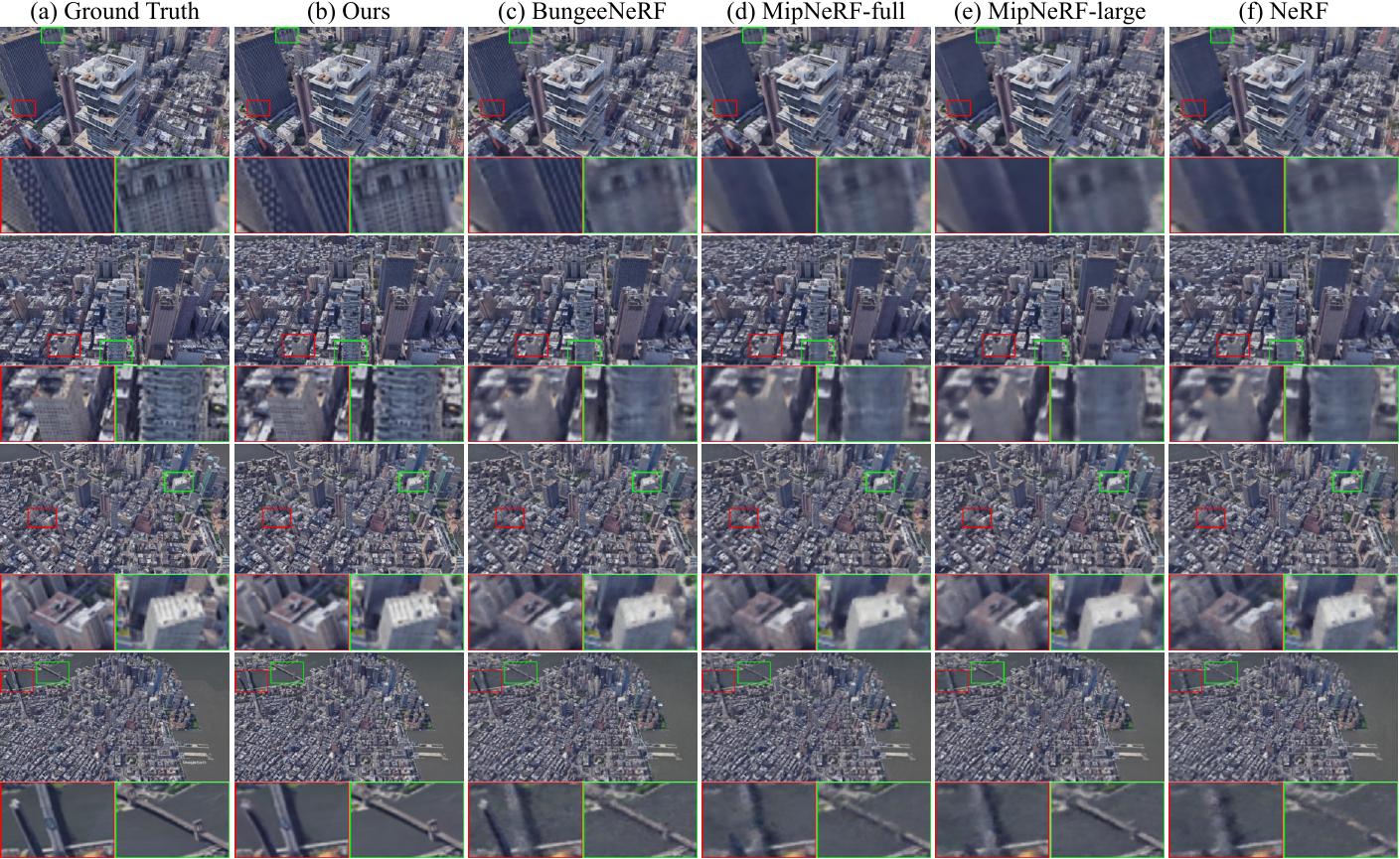}
	\caption{Qualitative comparisons on 56 Leonard dataset}
	\label{fig: Qualitative comparisons on 56Leonard}
\end{figure}

\begin{figure}[!h]
    \centering
    \includegraphics[width=\textwidth]{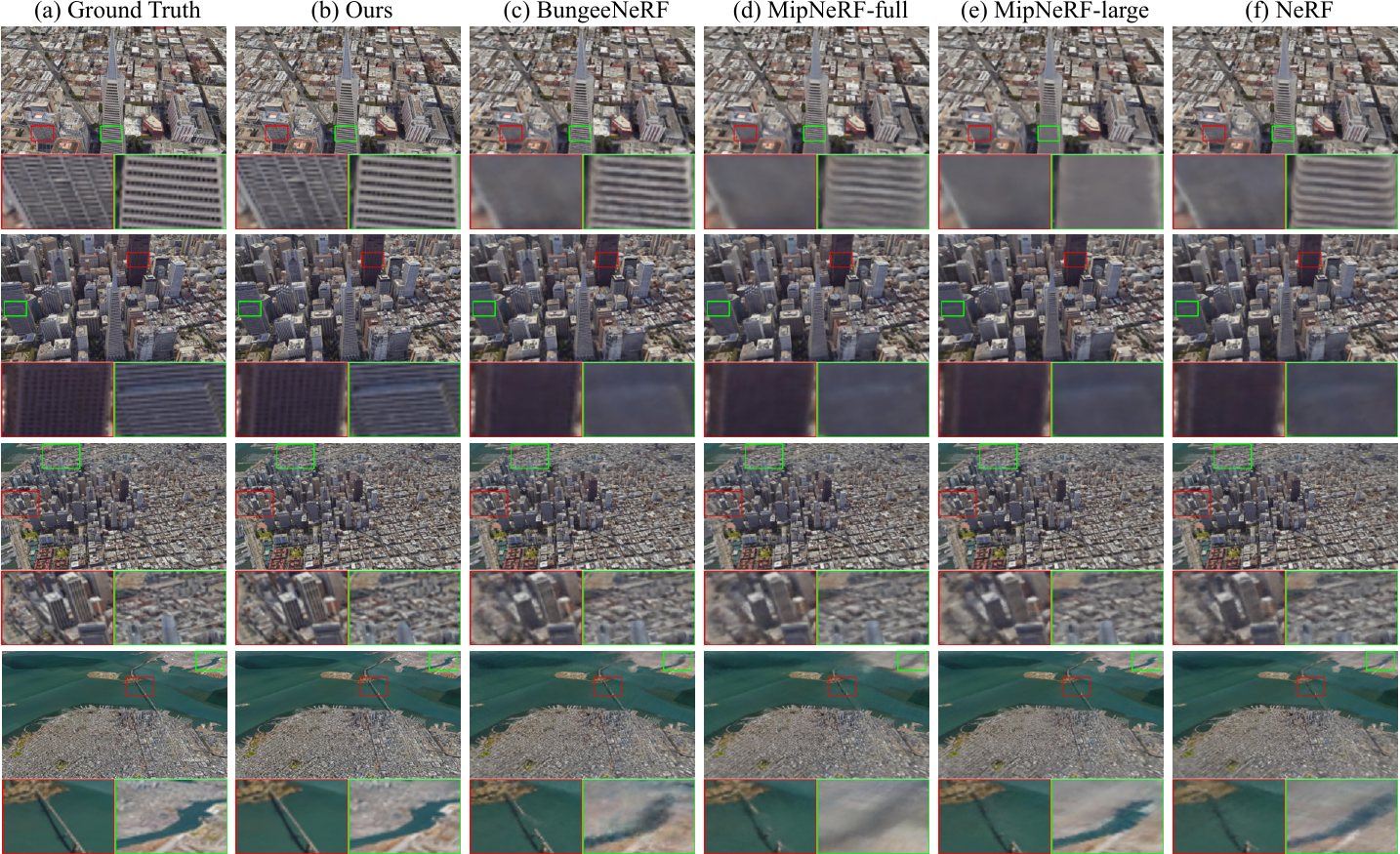}
	\caption{Qualitative comparisons on Transamerica dataset}
	\label{fig: Qualitative comparisons on Transamerica}
\end{figure}

\textbf{Training Details.}
Both coarse and fine points are set as ${N_c} = 64$ and ${N_f} = 64$ per ray. The image feature extraction network using a U-Net-like architecture with ResNet34 \cite{he2016deep} truncated after layer3 as the encoder, and two additional up-sampling layers with convolutions and skip-connections as the decoder. The dimension of the feature vector ${{f_i}\left( q \right)}$ is $35$. The number of source images is empirically set as $V = 10$. The Feature fusion module employs $2$ layers transformer block. The $ML{P_1}$ has $4$ layers with $64$ feature dimensions. The $ML{P_2}$ has only $1$ layer with $64$ feature dimensions. We apply the Adam as optimizer, with initial learning rates of $10{e^{ - 3}}$ for the feature extraction network and $5 \times 10{e^{ - 4}}$ for our tiny MLP. All experiments are conducted in $200k$ iterations with $2048$ rays sampled in each iteration on a single RTX $4090$ GPU.

\begin{table}[!h]
    \centering
    \caption{Quantitative comparisons on $56$ Leonard and Transamerica datasets. $D$ denotes $ML{P_1}$ depth, $d$ denotes $ML{P_1}$ width and $skip$ indicates which layer(s) the skip connection is inserted to.}
    \label{tab: Quantitative comparison}
    \renewcommand{\arraystretch}{1.3}
    \resizebox{\textwidth}{!}{
        \begin{tabular}{lccccccccc}
		\hline
		\multicolumn{4}{c}{} & \multicolumn{3}{c}{$56$ Leonard} & \multicolumn{3}{c}{Transamerica}\\
		\multicolumn{4}{l}{} & \multicolumn{1}{c}{PSNR $\uparrow$} & \multicolumn{1}{c}{SSIM $\uparrow$} & \multicolumn{1}{c}{LPIPS $\downarrow$} & \multicolumn{1}{c}{PSNR $\uparrow$} & \multicolumn{1}{c}{SSIM $\uparrow$} & \multicolumn{1}{c}{LPIPS $\downarrow$}\\
		\hline
            \multicolumn{4}{l}{NeRF \cite{nerf} ($D=8$, $d=256$, $skip=4$)} & \multicolumn{1}{c}{$23.570$} & \multicolumn{1}{c}{$0.729$} & \multicolumn{1}{c}{$0.324$} & \multicolumn{1}{c}{$24.046$} & \multicolumn{1}{c}{$0.747$} & \multicolumn{1}{c}{$0.311$}\\
            
            \multicolumn{4}{l}{Mip-NeRF-small \cite{barron2021mip} ($D=8$, $d=256$, $skip=4$)} & \multicolumn{1}{c}{$23.337$} & \multicolumn{1}{c}{$0.709$} & \multicolumn{1}{c}{$0.354$} & \multicolumn{1}{c}{$23.929$} & \multicolumn{1}{c}{$0.733$} & \multicolumn{1}{c}{$0.330$}\\
            
            \multicolumn{4}{l}{Mip-NeRF-large \cite{barron2021mip} ($D=10$, $d=256$, $skip=4$)} & \multicolumn{1}{c}{$23.507$} & \multicolumn{1}{c}{$0.718$} & \multicolumn{1}{c}{$0.346$} & \multicolumn{1}{c}{$24.113$} & \multicolumn{1}{c}{$0.744$} & \multicolumn{1}{c}{$0.315$}\\
            
            \multicolumn{4}{l}{Mip-NeRF-full \cite{barron2021mip} ($D=10$, $d=256$, $skip=4,6,8$)} & \multicolumn{1}{c}{$23.665$} & \multicolumn{1}{c}{$0.732$} & \multicolumn{1}{c}{$0.328$} & \multicolumn{1}{c}{$24.113$} & \multicolumn{1}{c}{$0.748$} & \multicolumn{1}{c}{$0.314$}\\
            
		\multicolumn{4}{l}{BungeeNeRF \cite{xiangli2022bungeenerf} ($D=10$, $d=256$, $skip=4,6,8$)} & \multicolumn{1}{c}{$24.513$} & \multicolumn{1}{c}{$0.815$} & \multicolumn{1}{c}{$0.160$} & \multicolumn{1}{c}{$24.415$} & \multicolumn{1}{c}{$0.801$} & \multicolumn{1}{c}{$0.192$}\\
  
            \multicolumn{4}{l}{Ours ($4k$ iterations, $D=4$, $d=64$, $skip=1$)} & \multicolumn{1}{c}{$24.570$} & \multicolumn{1}{c}{$0.837$} & \multicolumn{1}{c}{$0.259$} & \multicolumn{1}{c}{$25.063$} & \multicolumn{1}{c}{$0.846$} & \multicolumn{1}{c}{$0.262$}\\
            
		\multicolumn{4}{l}{Ours (same iter as baselines, $D=4$, $d=64$, $skip=1$)} & \multicolumn{1}{c}{\bm{$30.450$}} & \multicolumn{1}{c}{\bm{$0.963$}} & \multicolumn{1}{c}{\bm{$0.065$}} & \multicolumn{1}{c}{\bm{$30.599$}} & \multicolumn{1}{c}{\bm{$0.959$}} & \multicolumn{1}{c}{\bm{$0.076$}}\\
  
		\hline
	\end{tabular}
    }
\end{table}

\subsection{Experiment Results}
We compare the proposed AG-NeRF against the state-of-the-art method BungeeNeRF \cite{xiangli2022bungeenerf}, vanilla NeRF \cite{nerf}, and three forms of MipNeRF \cite{barron2021mip}.

We report the performance of our method and baselines in Fig. \ref{fig: Qualitative comparisons on Transamerica dataset and traintime}, \ref{fig: Qualitative comparisons on 56Leonard}, \ref{fig: Qualitative comparisons on Transamerica} and Table \ref{tab: Quantitative comparison}. It can be observed that there are significant improvements both in qualitative and quantitative metrics. In comparison with the SOTA method BungeeNeRF \cite{xiangli2022bungeenerf}, the proposed AG-NeRF achieves an increase of about $6$ dB in PSNR on both the $56$ Leonard and Transamerica datasets. In close views, baseline methods result in blurry textures due to substantial variation in detail levels, while the proposal can reconstruct sharper geometric and more delicate details. On the other hand, these competitors usually fail to reconstruct complete geometry in remote views, such as slender bridges or densely populated buildings located far from the camera. The proposed AG-NeRF can answer in affirmative to synthesize complete images.

The BungeeNeRF \cite{xiangli2022bungeenerf} is time-consuming because of its multi-stage training process. Each stage in this sequential training paradigm relies on the checkpoint file generated in the preceding stage, which means each stage has to be fully trained before progressing to the next. It takes up to $5$ days to finish training four stages of BungeeNeRF via a single RTX $4090$ GPU. In contrast, the proposed method only costs $1$ day to converge. 

It is worth noting that, with a distinct advantage in leveraging scene priors from our source images, the proposal AG-NeRF facilitates the rapid acquisition of overall geometry of scenes and then gradually recovers texture details. As evidenced on the right of Fig. \ref{fig: Qualitative comparisons on Transamerica dataset and traintime}, our model, with shallower MLP, using only $4k$ iterations of training (about a half hour) can achieve competitive performance compared to BungeeNeRF.

\begin{table}[h]
    \centering
    \caption{Ablation study. When the number of source images $V=0$, the framework is equivalent to the vanilla NeRF ($D=4$, $d=64$, $skip=1$). No attention (AvgPool) means avg-pooling the interpolated feature vectors directly. No attention (MaxPool) means max-pooling the interpolated feature vectors directly.}
    \label{tab: Ablation study on the attention-based feature fusion strategy}
    \renewcommand{\arraystretch}{1.3}
    \resizebox{\textwidth}{!}{
        \begin{tabular}{lcccccccc}
		\hline
		\multicolumn{2}{c}{} & \multicolumn{3}{c}{$56$ Leonard} & \multicolumn{3}{c}{Transamerica}\\
		\multicolumn{2}{c}{} & \multicolumn{1}{c}{PSNR $\uparrow$} & \multicolumn{1}{c}{SSIM $\uparrow$} & \multicolumn{1}{c}{LPIPS $\downarrow$} & \multicolumn{1}{c}{PSNR $\uparrow$} & \multicolumn{1}{c}{SSIM $\uparrow$} & \multicolumn{1}{c}{LPIPS $\downarrow$}\\
		\hline
  
  	\multicolumn{2}{l}{No source image ($V=0$)} & \multicolumn{1}{c}{$19.250$} & \multicolumn{1}{c}{$0.370$} & \multicolumn{1}{c}{$0.608$} & \multicolumn{1}{c}{$20.027$} & \multicolumn{1}{c}{$0.416$} & \multicolumn{1}{c}{$0.588$}\\
   
		\multicolumn{2}{l}{No attention (AvgPool)} & \multicolumn{1}{c}{$28.719$} & \multicolumn{1}{c}{$0.941$} & \multicolumn{1}{c}{$0.101$} & \multicolumn{1}{c}{$28.743$} & \multicolumn{1}{c}{$0.935$} & \multicolumn{1}{c}{$0.120$}\\
  
		\multicolumn{2}{l}{No attention (MaxPool)} & \multicolumn{1}{c}{$28.994$} & \multicolumn{1}{c}{$0.946$} & \multicolumn{1}{c}{$0.090$} & \multicolumn{1}{c}{$28.950$} & \multicolumn{1}{c}{$0.939$} & \multicolumn{1}{c}{$0.109$}\\
   
		\multicolumn{2}{l}{Ours} & \multicolumn{1}{c}{\bm{$30.450$}} & \multicolumn{1}{c}{\bm{$0.963$}} & \multicolumn{1}{c}{\bm{$0.065$}} & \multicolumn{1}{c}{\bm{$30.599$}} & \multicolumn{1}{c}{\bm{$0.959$}} & \multicolumn{1}{c}{\bm{$0.076$}}\\
  
		\hline
	\end{tabular}
    }

\end{table}

\begin{figure}[h]
    \centering
    \includegraphics[width=\textwidth]{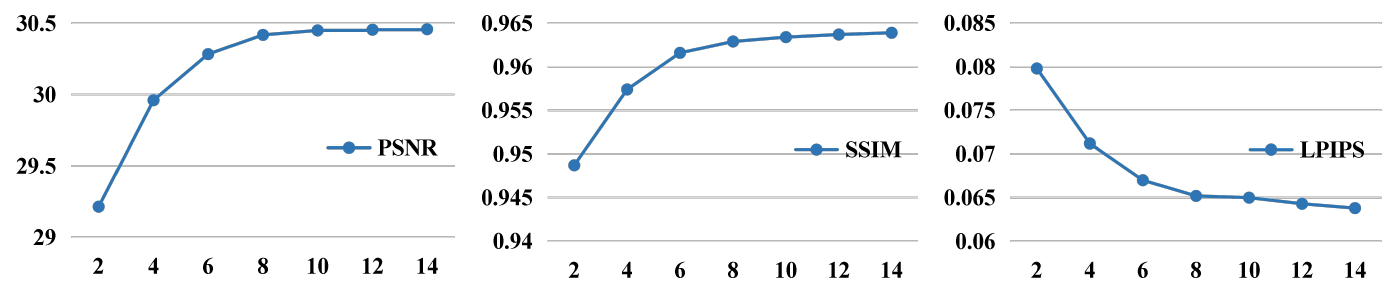}
	\caption{Comparison on the effect of source image number. The horizontal axis represents the number of chosen source images.}
	\label{fig: Ablation study on the number of source images}
\end{figure}

\subsection{Ablation Study}
\textbf{Effectiveness of the Source Images.} To validate the effectiveness of source images to provide scene priors, we test our approach on the $56$ Leonard dataset using varying numbers of source images. Fig. \ref{fig: Ablation study on the number of source images} shows that, as the number of source images increases from $2$ to $14$, our method achieves about $1.3$ dB gain in PSNR on the $56$ Leonard dataset. When the number of source images exceeds $10$, there is a tiny margin on $3$D reconstruction result. Therefore, to balance the trade-off between computational efficiency and rendering quality, we choose $10$ source images for experiments. 

Furthermore, we conduct another ablation study excluding source image selection method ($V = 0$), that is, vanilla NeRF \cite{nerf}. This is because the attention-based feature extraction and fusion module fails to take effect without the inputs of source images. The results are summarized in the first row of Table \ref{tab: Ablation study on the attention-based feature fusion strategy}, compared to ours, there is about $11$ dB decrease in PSNR.

\textbf{Effectiveness of Attention-based Feature Fusion Approach.} To evaluate the effectiveness of our attention-based feature fusion approach in capturing features corresponding to target pixel, avg-pooling or max-pooling are applied for the interpolated feature vectors. As shown from the second and third rows of Table \ref{tab: Ablation study on the attention-based feature fusion strategy}, under the same number of source images, avg-pooling case results in an approximately $1.7$ dB decrease in PSNR, and max-pooling case leads to about $1.5$ dB decrease in PSNR as compared to our model.

\section{Conclusion}
In this work, we target on rendering remarkable large-scale outdoor scenes captured at varying altitudes. To that end, we propose an end-to-end framework, termed AG-NeRF. NeRF-based approaches  typically require hours to days to be trained. Advanced NeRF variants can alleviate this, but tend to be less accessible to the large-scale outdoor scene reconstruction due to the requirement of low-level rendering, complex parameter tuning or a computational bottleneck.  It is envisioned that the proposal can model diverse multi-scale scenes with drastically varying city views. Experiments demonstrate that AG-NeRF is competitive to the SOTA in terms of accuracy and speed.

\bibliographystyle{IEEEbib}
\bibliography{ref}

\end{document}